# Supporting Energy Policy Research with Large Language Models

*Grant Buster*[*1], *Pavlo Pinchuk*[1], *Jacob Barrons*[1], *Ryan McKeever*[1], *Aaron Levine*[1], *Anthony Lopez*[1]

*Corresponding Author: Grant.Buster@nrel.gov*

[1] *National Renewable Energy Laboratory, 15013 Denver W Pkwy, Golden, CO 80401, US*

## Abstract

The recent growth in renewable energy development in the United States has been accompanied by a simultaneous surge in renewable energy siting ordinances. These zoning laws play a critical role in dictating the placement of wind and solar resources that are critical for achieving low-carbon energy futures. In this context, efficient access to and management of siting ordinance data becomes imperative. The National Renewable Energy Laboratory (NREL) recently introduced a public wind and solar siting database to fill this need. This paper presents a method for harnessing Large Language Models (LLMs) to automate the extraction of these siting ordinances from legal documents, enabling this database to maintain accurate up-to-date information in the rapidly changing energy policy landscape. A novel contribution of this research is the integration of a decision tree framework with LLMs. Our results show that this approach is 85 to 90% accurate with outputs that can be used directly in downstream quantitative modeling. We discuss opportunities to use this work to support similar large-scale policy research in the energy sector. By unlocking new efficiencies in the extraction and analysis of legal documents using LLMs, this study enables a path forward for automated large-scale energy policy research.

## 1  Introduction

The energy system is rapidly changing as are the markets and policies that govern it. In recent years, the United States has seen a rise in wind and solar renewable energy generation [1,2]. In response, electricity markets have introduced new market structures and local governments have developed new regulations to accommodate these novel generators [3,4,5]. From 2018 to 2022, the electricity generation from utility-scale wind and solar sources has increased by 60 and 125 percent, respectively [2]. Over the same period, the number of local zoning ordinances governing the development of wind energy systems has increased by more than 540 percent (solar ordinances were not collected in 2018 so a percent change cannot be calculated) [5]. These policies directly affect the quantity of available, buildable land for renewables and have important implications for our future energy system [5].

These changes are happening at the same time the United States is setting ambitious goals for clean energy deployment in an effort to tackle the climate crisis [6]. Several scientific institutions have recently charted technical paths to achieve such decarbonization goals while identifying key opportunities and challenges [7,8]. However, without accurate data on current local energy policies, these analyses may represent idealized energy transitions without a realistic representation of the local legal challenges to deployment of renewable energy assets. Similar developments are entirely possible in numerous other legal, economic, and political domains that surround the energy industry.

In the context of these changes, efficient access to and management of policy data becomes crucial. Historically, the process of analyzing energy policy has only been possible through significant manual labor. For example, the United States Wind and Solar Siting Regulation and Zoning Ordinance Databases recently introduced by the National Renewable Energy Laboratory (NREL) [5,9,10] required approximately 1,500 human labor hours to collect siting regulation data from local zoning codes in the United States. This significant labor requirement makes it prohibitively expensive to continually maintain accurate data in these dynamic times. However, technologies in machine learning, natural language processing (NLP), and semantic search have also seen transformative developments in recent years, promising to enable the automation of these tasks.

The advent of Large Language Models (LLMs) such as GPT-4 has changed our understanding of what is possible in automated reading comprehension and text processing [11,12]. Applications of LLMs are being explored in medicine, law, education, finance, engineering, and media [13]. Accordingly, there are numerous possibilities for LLMs to expedite the retrieval and processing of energy policy data from legal documents. We see this as an ideal testing ground for real-world applications of LLMs because of the substantial reading comprehension requirements and the otherwise prohibitive cost of maintaining up-to-date information on the current state of energy policies across the nation.

In this work, we demonstrate how LLMs can be used to retrieve data on renewable energy zoning ordinances from legal documents. We introduce a strategy using decision trees to supplement the LLM fast reasoning skills with subject matter expertise and symbolic logic that leads to improved accuracy when compared to human effort or alternative LLM prompting strategies. Finally, we discuss future applications of the open-source software used here to enable similar large-scale data retrieval efforts elsewhere.

The structure of this article is as follows: Section 2 introduces the experimental methods along with the LLM prompting strategies. Section 3 discusses the results of applying LLMs to a set of ordinance documents. Section 4 discusses the potential for future opportunities and limitations of these methods in energy research. Section 5 concludes the article.

## 2 Methods

To explore the capability of LLMs to assist in the retrieval of renewable energy siting data from local ordinance documents, we collect a subset of legal documents from counties represented in the United States Wind Siting Regulation and Zoning Ordinances Database (hereafter referred to as "the wind ordinance database") [9].

We split the documents into two sets: 18 documents in the "training" set and 85 documents in the "test" set. Each document here represents ordinances for a single county with known restrictions on wind energy systems. Although the LLM framework is not trained through the direct observation of numerical data like a traditional quantitative regression model might be, we find a similar training and test experimental setup useful to ensure extensibility of our approach. Here, the documents in the training set are used to develop the LLM approach. This includes development of the software for text cleaning and the LLM prompting strategies. The software and prompting strategies are then frozen and run against the test set documents. The test set documents are a "ground truth" dataset compiled by two researchers not involved in the development of the LLM strategies. The two researchers reviewed all 85 documents separately, resolved any discrepancies, and produced the most accurate test dataset possible. Comparable performance of the LLM strategies in both the training and test sets supports the extensibility of these methods, especially when considering the heterogeneity of ordinance text in the documents.

The document parsing methods are split into three parts. First, ordinance documents typically in Portable Document Format (PDF) are converted to text. Second, the ordinance text is reduced to only the text that is relevant to wind energy systems. Lastly, the LLM is used to extract data from the text.

The first part of the methods requires conversion of legal documents from PDF to text. We explored several utilities for PDF-to-text conversion and found that the free software Poppler [14] performed best, especially when considering the ability to convert ordinances in tabular format to space-delimited text.

The second part of the methods requires distilling the ordinance text to only the text relevant to wind energy systems. Legal documents on zoning can be hundreds of pages long, easily exceeding the context window of most publicly available LLMs, including GPT-4 (4k or 32k tokens at the time of development). Although the context window of publicly available LLMs is rapidly increasing, long-window models can still suffer from low accuracy when requested to extract information from the middle of a large document [15]. As a result, we found that content distillation was an important step to improve the accuracy of these methods.

In order to distill the lengthy legal documents, we first perform an asynchronous semantic search on overlapping chunks of the text, instructing the LLM to "extract text related to the restrictions of wind energy systems" (see the code associated with this paper for full details on LLM prompts). A challenge here is preserving parts of the document that do not explicitly define the ordinances but that set important context such as definition of terms. Generally, however, a 100-page document on all zoning ordinances in a county typically does not have more than a few pages relevant to wind energy systems. This step distills the relevant information into a manageable amount of text that can be used by the LLM. An example of the desired output for a few select counties can be seen in Table 1.

Table 1. Sample text on wind energy siting from ordinance documents.

| County | Pages | Sampled Text |
|---|---|---|
| Laramie, WY | 223 | The center of the base of each wind tower shall be located no less than 1.5 (hub height + rotor diameter) from adjacent unplatted nonparticipating property lines and dedicated public roads. |
| Ottowa, MI | 16 | Medium Wind Energy Turbine (MWET) shall also be subject to the following:<br>• Occupied Building Setback: The setback from all occupied buildings on the applicant's parcel shall be a minimum of twenty (20) feet measured from the base of the Tower.<br>Large Wind Energy Turbine (LWET) shall also be subject to the following:<br>• Occupied Building Setback: Each LWET shall be set back from the nearest Occupied Building that is located on the same parcel as the LWET a minimum of two (2) times its Total Height, or one thousand (1000) feet, as measured from the base of the Tower, whichever is greater. |
| Monroe, WI | 26 | Occupied community buildings:<br>• The lesser of 1,250 feet or 3.1 times the maximum blade tip height.<br>Participating residences:<br>• 1.1 times the maximum blade tip height.<br>Nonparticipating residences:<br>• The lesser of 1,250 feet or 3.1 times the maximum blade tip height |

The last part of the method is to extract the discrete siting restrictions from the legal text relevant only to wind energy systems. The goal here is to output quantitative data that can be used in analysis such as that by Lopez et al., 2022 [9]. For this study, we chose 13 features that wind energy systems are commonly required to consider in siting decisions: structures (participating and non-participating), property lines (participating and non-participating), roads, railroads, transmission lines, bodies of water, noise restrictions, maximum system heights, minimum lot sizes, shadow flicker restrictions, and turbine density restrictions. The siting ordinances relative to the physical features are typically called setbacks (e.g., a turbine must be "set back" a pre-defined distance from these features). For this process to be useful in downstream quantitative analysis, the output must be concise and machine readable. An example query might be "what is the setback from participating residences in Monroe, Wisconsin?" and the output should consist of a numerical value and a categorical definition e.g., the value would be 1.1 and the categorical definition would be "maximum tip height multiplier". In this example, we know the county and the feature (Monroe; participating residences) and we can extract the numeric setback value and how to calculate the final distance value (1.1; a multiple of the maximum tip height). While this may be an intuitive task for a human, the small set of ordinance text in Table 1 shows the potential for significant heterogeneity in how these setbacks are defined in the real world. There are frequently multiple setback values to choose from, multiple subclasses of feature within a broader feature category, multiple classes of turbine based on size and application, and a wide variety of verbiage and abbreviations.

The strategy we found to work best for extracting ordinance data from legal documents is an approach using LLMs guided by a decision tree embedded with knowledge of the subject matter. Using a decision tree as a programmatic structure, we can break down our larger goal into many smaller requests (nodes) with logical transitions (edges) between each request (see Figure 1). This is essentially a pre-programmed multi-prompt LLM conversation based on subject matter expertise. In practice, this looks like a series of nodes each containing an LLM prompt ("does the text define a setback?") with functions ("is the answer yes?") determining transitions between nodes. We also implement branching logic to handle the various types of ordinances that are recorded in the original wind ordinance database, including branches for ordinances with multiple conditions and instructions on what assumptions to make under this ambiguity (see Figure 2). The output of this structure can be the setback text in our desired format (including as structured data) as returned by a leaf node (e.g., a node with no edges directed away from it) or an exception raised because no edge transition conditions were met. In the latter case, depending on where in the tree the exception was raised, we can assume there is no setback defined or flag the ordinance for human review. This approach is slower than a single-prompt strategy because it uses multiple chained prompts per document but is similarly limited only by the rate limit imposed by the LLM provider. A description of the decision tree algorithm is documented below in Algorithm 1.

This implementation of a decision tree to guide LLMs represents a rudimentary implementation of neuro-symbolic artificial intelligence [16], combining the powerful text-processing capabilities of the neural network with the ability for a subject matter expert to craft a logical structure for how a human would extract relevant information from a document.

The decision tree strategy is also similar to the tree-of-through approach [17,18] but has some fundamental differences. Most crucially, in our data retrieval task we do not rely on self-evaluation of progress by an LLM or a rule-based algorithm. Instead, we rely on human-developed instruction and logic to guide the edge transitions and raise exceptions in the case of unrecognized ordinance types. Also, in our use case we define only a directed acyclic graph, whereas the tree-of-thought approach allows for backtracking through the graph. The software used here can accommodate cyclic graphs, but the acyclic assumption is convenient for this use case and prevents infinite loops.

It is also worth noting that the software we use to implement the decision tree is sufficiently generic to allow for any callable function to serve as a node or edge property. More complex problems could easily implement complex logical functions, neural networks, or nested graphs to processes edge conditions.

We also ran experiments with a Retrieval-Augmented Generation (RAG) framework [19] and single-prompt data retrievals instead of the multi-prompt decision tree. We found that both strategies were ineffective for retrieving accurate structured information from a larger body of legal text. The heterogeneity of ordinance verbiage and setback values in tabular structures rendered vector comparison in a RAG framework unreliable. The complexity of the retrieval query also proved to be too challenging for a single large prompt, especially when the LLM was faced with ambiguous conflicting information such as with regulations on both small and large wind energy systems or with regulations that required comparative logic between multiple values. Both experiments were also unable to provide output in consistently machine-readable quantitative values paired with discrete categories for the ordinance type (the ability of decision trees to do this is discussed later in Section 3). In theory, we could augment single-prompt approaches with additional strategies such as providing the LLM with an example of successful ordinance extraction to guide its thought process [20] but given the diversity of possible ordinance examples we decided this was not tractable. Ultimately, we did not run these strategies on the full test set of ordinance documents because of poor performance in development with the training set of documents.

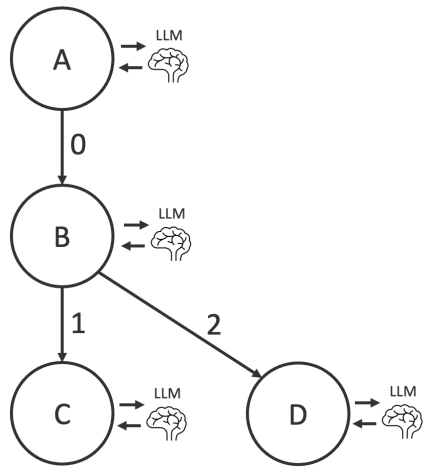

**Figure 1. Example architecture of the decision tree approach. Nodes A-D interact with the LLM using unique prompts and the conversation history. Edges 0-2 each have callable functions or additional LLM prompts that would trigger node transitions. Failure of all edge conditions in response to an LLM output at a node results in a null result (e.g., no relevant text) or human review. Leaf nodes C and D result in usable outputs based on their unique conditions. For example, leaf node C could return a wind turbine tip height multiplier setback while node D could return a fixed distance setback.**

**Algorithm 1. Decision tree pseudocode.**

```
1   Input: body of text from the document in question t, target setback feature f
2   Function DecisionTree.run(t, f):
3       node = DecisionTree.InitialNode
4       while is_not_leaf_node(node):
5           prompt = Prompter(t, f, node)
6           response = LLM(prompt)
7           Edges = GetEdges(node)
8           for Edge_i in Edges:
9               condition = Edge_i(response)
10              if condition:
11                  node = get_next_node(Edge_i)
12                  break
13          if not condition:
14              raise Exception(prompt, node, response)
15      return response
```

# 3 Results

Table 2 and Table 3 summarize the results of the analysis on the training and test sets, respectively. The decision tree LLM strategy was found to be between 85% and 90% accurate between the two sets of documents. Additionally, the decision tree outputs are structured in machine-readable format and can be used directly in downstream quantitative analysis. The comparable accuracy between the training and test documents supports the extensibility of this strategy to a larger set of ordinance documents.

**Table 2. Results of ordinance retrieval from 18 training documents with 13 ordinance types (234 total ordinance values). Overall accuracy, precision, and recall were 86%, 96%, and 75% respectively.**

|  |  | LLM Found Ordinance | |
|---|---|---|---|
|  |  | True | False |
| Ordinance Exists | True | 61 (26%) – Correct<br>11 (5%) – Incorrect | 20 (9%) |
|  | False | 2 (1%) | 140 (60%) |

**Table 3. Results of ordinance retrieval from 85 test documents with 13 ordinance types (1105 total ordinance values). Overall accuracy, precision, and recall were 90%, 91%, and 81% respectively.**

|  |  | LLM Found Ordinance | |
|---|---|---|---|
|  |  | True | False |
| Ordinance Exists | True | 325 (29%) – Correct<br>34 (3%) – Incorrect | 42 (4%) |
|  | False | 34 (3%) | 670 (61%) |

These results are comparable to small-scale (around 20 counties) internal validation showing that the original human-developed wind ordinance database is also between 85% and 90% accurate. This result is understandable when considering the immense number of human labor hours required for the initial development of the wind ordinance database (1,500 hours), the repetitive nature of the work required, and the repeated identification of fatigue as a common driver in human errors [21]. In fact, even in the development of the 85-document test set, ambiguity and fatigue led to initial disagreement between the two researchers tasked with development of the "ground truth" dataset. This was resolved upon comparison of results from the two researchers, but it highlights the difficulty of this problem.

An example of the decision tree strategy extracting the structure setback data from the Monroe, Wisconsin ordinance document is shown in Figure 2. This example is highlighted because it requires some of the most complex mathematical and logical reasoning found in ordinance documents. The ordinance states different setback requirements from three types of structures, with some types of structures having multiple options to calculate and compare. The LLM shows its work while calculating all possible effective setback distances and is able to select an appropriate final value based on the guidance provided. In contrast, experiments with a single-prompt LLM containing the same guidance returned "the setback is 3.1 times the maximum blade tip height" without showing its work. This is a reasonable output, but it is incorrect given the provided guidance.

> **USER**: Is there text in the legal document that describes how close i can site or how far i have to setback wind energy systems to buildings, structures, and/or residences? Please only say "Yes" or "No".
>
> **ASSISTANT**: Yes
>
> **USER**: Based on your last message, choose the option that best describes the required setback from buildings, structures, and/or residences:
>     #1.1 The setback from buildings and structures is a single distance value
>     #7.1 The setback from buildings and structures has multiple conditions such as a fixed distance and a multiple of the total system height
>     #0 None of the above options are descriptive of this ordinance related to setbacks from buildings and structures
>
> **ASSISTANT**: #7.1 The setback from buildings and structures has multiple conditions such as a fixed distance and a multiple of the total system height
>
> **USER**: Lets assume the supporting tower is 377 feet with blades 279 feet long. So the rotor diameter would be 558 feet and the total system height would be 656 feet. What would the multiple setback distances be? Please show your work.
>
> **ASSISTANT**: Based on the information provided, we can calculate the setback distances for the different types of buildings and structures:
>     1. Occupied community buildings: The lesser of 1,250 feet or 3.1 times the maximum blade tip height 3.1 * 656 feet = 2,033.6 feet. The lesser value is 1,250 feet.
>     2. Participating residences: 1.1 times the maximum blade tip height 1.1 * 656 feet = 721.6 feet
>     3. Nonparticipating residences: The lesser of 1,250 feet or 3.1 times the maximum blade tip height 3.1 * 656 feet = 2,033.6 feet. The lesser value is 1,250 feet.
> So, the multiple setback distances are:
>     1,250 feet for occupied community buildings
>     721.6 feet for participating residences
>     1,250 feet for nonparticipating residences
>
> **USER**: What is the final setback value? If the ordinance states which of the multiple distances should be chosen, use that guidance. Otherwise, choose the largest setback value.
>
> **ASSISTANT**: The ordinance does not provide specific guidance on choosing between the multiple distances. Therefore, we will choose the largest setback value. The final setback value is 1,250 feet.
>
> **USER**: State the final setback like this: "The setback is XXX (units)"
>
> **ASSISTANT**: The setback is 1,250 feet.

**Figure 2. Example of decision tree path for a complex setback rule from the Monroe, Wisconsin Code of Ordinances. Note that for illustrative purposes we only abridged messages from the full conversation that occurred during the experiment.**

An additional benefit of the decision tree method not represented in the accuracy metrics is the ability for the decision to act as a classification model in addition to a data extraction utility. One node of the decision tree requests the LLM to categorize the ordinance based on a set of archetypes from the original ordinance database (e.g., tip-height multiplier, fixed distance value, etc…). The result is that the type of setback can be easily determined from the decision tree history and can be output in a machine-readable format. This enables the scalability of this method to the full wind ordinance database.

As with any machine learning model applied to the real world, it is crucial to understand how the model might fail and what the consequences might be. In the training and test results, we categorize the failures into three modes discussed further below.

The first failure mode is a false positive where the LLM identifies a regulation that is not present. In our initial tests we found that the LLM can "hallucinate" fictional regulations. This was primarily during experiments using the LLM to reformat massive bodies of text. The LLM appeared to perform poorly when requested to output a large body of text that was a highly faithful reproduction of the input prompt. This problem was mostly eliminated by explicitly instructing the LLM to not add additional text, by only using the LLM to extract small chunks of relevant text from the larger document, and by implementing a heuristic N-Gram similarity check [22] to further prevent large-scale hallucinations. In the final experiments, this failure mode occurred between 1 and 3% and appeared to be primarily due to the LLM mistaking a real regulation as applying to an erroneous feature (e.g., assuming that a setback applied to structures and property lines even though it only applied to structures).

The second failure mode is a false negative where the model fails to retrieve an ordinance that was present in a document. A major cause of this error type was when the PDF conversion process outputted poorly formatted text. The LLM also sometimes failed to recognize text as relevant to the question, although this was rare. In training, this was the most common error mode occurring about 9% of the time, almost double each of the other two failure modes. However, when applying the LLM to the larger test set, the false negative rate decreased to 4% which was only slightly more common than the other two failure modes.

The last failure mode occurred when the LLM found an ordinance, but it was not the correct ordinance for the request. As an example, many ordinance documents specify regulations for different types of residences and for different sizes of turbines, which can sometimes confuse the LLM especially when the text formatting is not intuitive. Anecdotally, there were also several cases where researchers did not initially agree on the correct interpretation of the ordinance text, so clearly this failure mode is not limited solely to LLMs.

In all failure modes, the frequency and consequence of such errors is relatively small with respect to these specific research outputs and their application to large-scale energy analysis. Energy developers should still work with licensed attorneys to ensure their projects adhere to local ordinances. This is discussed further in the next section.

# 4 Discussion

The results of this paper support the use of LLMs in the continued maintenance of national-scale wind siting ordinances. However, the software used in this work may also enable similar national-scale data retrieval for analysis of utility rate structures [23], challenges in transmission siting [24], and other impacts of state-to-state variation in socio-political context [25]. The following paragraphs discuss the applicability and limitations of such strategies with respect to the lessons learned from the work presented here.

First, the challenge presented by maintaining a national database of wind siting ordinances is an ideal task for automation by LLMs because it is based on reading comprehension that has historically required significant human effort and is prone to errors from human fatigue.

Although LLMs have been developing rudimentary data analysis skills, their primary value proposition is still rooted in text-based reading comprehension. That is why we focus on the application of LLMs to policy data, which is typically stored in a heterogeneous collection of text.

Second, applications of LLMs are discussed here as tools for policy research and not as certified legal counsel. LLM outputs are best used in applications that are not safety-critical and have no risk asymmetry. That is, even if we engineer a process in which LLMs can be highly accurate, we should always ensure that inaccurate LLM responses will not result in severe real-world consequences. Users of the ordinance database should take steps to minimize the impacts of inaccurate values, and developers designing a wind plant should still consult the local zoning codes with help from the appropriate legal counsel. The value of such national-scale data comes from the ability to perform comparative analysis across regions, to maintain an understanding of national trends in renewable energy policy, to identify opportunities for effective future renewable energy policies, and to act as an initial screening tool for development opportunities. The distributed nature of the problem is important to highlight here: a single incorrect ordinance resulting from an inaccurate LLM response (or from human effort) will not invalidate the larger research and analysis effort.

Thirdly, the application of LLMs to zoning ordinance documents presents no data privacy or intellectual property challenges. All policies analyzed in this document are publicly available and contain no sensitive information. It is still somewhat unclear how text submitted in LLM prompts may be used by LLM providers, and policies vary from provider to provider (e.g., OpenAI and Google may have different policies on data privacy with respect to their LLMs). Applying LLMs to sensitive information may be possible, but such applications add additional challenges to the application of this new technology.

Lastly, we believe that great care should be taken in the application of LLMs to any real-world challenge. We describe our experimental setup in Section 2 for testing the LLM strategies on documents that are hidden from the primary developer. We plan to implement a sampled validation procedure when working with the full set of ordinance documents and recommend similar validation exercises be performed when applying LLMs to new problems. These models are black boxes with little transparency behind their decision-making processes even when asked to show their work. Applying an LLM as an analysis tool to a pre-existing body of text reduces the likelihood that the model will generate fictional responses (e.g., "hallucinations"), but it is still possible. In this work, we found that we could greatly reduce the risk of fictional responses by stating that the model was only allowed to return text from the original legal document, by reducing the amount of text we expected the model to output, and by always explicitly providing an option for the model to provide a negative answer (e.g., "there is no such ordinance in the provided text").

# 5  Conclusions

The effort required to collect local policy data at a national-scale is considerable and many institutions do not have sufficient resources to successfully execute such work, even if the results would be of significant value to the energy industry and research community. The challenge of ensuring accurate information in a constantly changing industry presents an additional burden, which may incur similar costs but yield lesser benefits compared to initial endeavors.

In this work, we present strategies for the automated extraction of wind turbine setback data from ordinance documents. We show that an LLM guided by a decision tree framework can reliably extract siting ordinances from legal documents while requiring little human supervision or review. At a minimum, the methods presented here appear to be a viable path forward for maintaining accurate and up-to-date information on national wind siting ordinances. The results are intended primarily for research purposes and are no substitute for certified legal counsel when designing and developing a real renewable energy project. The true value proposition of this work is to enable high-quality policy analysis in a rapidly-evolving sector at an unprecedented scale.

Not explored here is the possibility for LLMs to enable retrieval of such documents from the internet via automated search. These methods are still in development and the expanding offerings from the private sector are also rapidly improving what is possible on this front. However, we are optimistic that LLMs can assist with the automation of a significant portion of this task, especially given the strong baseline of known ordinances in the original U.S. Wind and Solar Siting Regulation and Zoning Ordinances Databases [9,10].

# Abbreviations

| | |
|---|---|
| GPT | Generative Pre-trained Transformer |
| LLM | Large Language Model |
| LWET | Large Wind Energy Turbine |
| MWET | Medium Wind Energy Turbine |
| NLP | Natural Language Processing |
| NREL | National Renewable Energy Laboratory |
| PDF | Portable Document Format |
| RAG | Retrieval-Augmented Generation |

# Code and Data Availability

The software and data used in this work is open-source and available at no cost here: https://github.com/NREL/elm. The ordinance documents used in this report and the results from the experiments are available here: https://github.com/NREL/elm/tree/main/examples/ordinance_gpt. Note that GPT-4 application programming interface (API) was used in the development of this work. Specifically, the Microsoft Azure "2023-03-15-preview" deployment of GPT-4 was used. At the submission of this paper, the GPT-4 API is not freely available to the public and requires a paid subscription with OpenAI or Microsoft Azure.

# Acknowledgements

The authors would like to thank Keith Benes, Patrick Gilman, Dan Bilello, Meghan Mooney, and Taylor Curtis for their thoughtful reviews of an earlier draft.

This work was authored by the National Renewable Energy Laboratory, operated by Alliance for Sustainable Energy, LLC, for the U.S. Department of Energy (DOE) under Contract No. DE-AC36-08GO28308. Funding provided by the DOE Wind Energy Technologies Office (WETO), and internal research funds at the National Renewable Energy Laboratory. The views expressed in the article do not necessarily represent the views of the DOE or the U.S. Government. The U.S. Government retains and the publisher, by accepting the article for publication, acknowledges that the U.S. Government retains a nonexclusive, paid-up, irrevocable, worldwide license to publish or reproduce the published form of this work, or allow others to do so, for U.S. Government purposes.

## Disclosures

During the preparation of this work the authors used GPT-4 as a central piece of the experimental methods. After using this tool/service, the authors reviewed and edited the content as needed and take full responsibility for the content of the publication.